\documentclass[conference]{IEEEtran}
\IEEEoverridecommandlockouts
\usepackage{cite}
\usepackage{amsmath,amssymb,amsfonts}
\usepackage{algorithmic}
\usepackage{graphicx}
\usepackage{textcomp}
\usepackage{xcolor}
\usepackage{times}
\usepackage{latexsym}
\usepackage{booktabs}
\usepackage{xspace}
\usepackage{helvet}
\usepackage{tcolorbox}
\usepackage{enumitem}
\usepackage{verbatim}
\usepackage{anyfontsize}
\usepackage{wrapfig}
\usepackage{subcaption}
\usepackage{tabu}
\usepackage{algorithm}
\usepackage{multicol}
\usepackage{multirow}
\usepackage{listings}
\usepackage{colortbl}
\usepackage{url}
\definecolor{lightblue}{HTML}{F0F8FF}
\def\BibTeX{{\rm B\kern-.05em{\sc i\kern-.025em b}\kern-.08em
    T\kern-.1667em\lower.7ex\hbox{E}\kern-.125emX}}
\begin{document}

\title{Bridging Writing Manner Gap in Visual Instruction Tuning by Creating LLM-aligned Instructions\thanks{\textbf{This work was completed in November 2023.}}}

\author{Dong Jing, Nanyi Fei, Zhiwu Lu$^\dagger$\thanks{$^\dagger$ Corresponding Author} \\
Gaoling School of Artificial Intelligence, Renmin University of China \\
\texttt{\{jingdong98, luzhiwu\}@ruc.edu.cn}
}

\maketitle

\begin{abstract}
In the realm of Large Multi-modal Models (LMMs), the instruction quality during the visual instruction tuning stage significantly influences the performance of modality alignment. In this paper, we assess the instruction quality from a unique perspective termed \textbf{Writing Manner}, which encompasses the selection of vocabulary, grammar and sentence structure to convey specific semantics. We argue that there exists a substantial writing manner gap between the visual instructions and the base Large Language Models (LLMs) within LMMs. This gap forces the pre-trained base LLMs to deviate from their original writing styles, leading to capability degradation of both base LLMs and LMMs. To bridge the writing manner gap while preserving the original semantics, we propose directly leveraging the base LLM to align the writing manner of soft-format visual instructions with that of the base LLM itself, resulting in novel LLM-aligned instructions. The manual writing manner evaluation results demonstrate that our approach successfully minimizes the writing manner gap. By utilizing LLM-aligned instructions, the baseline models LLaVA-7B and QwenVL demonstrate enhanced resistance to hallucinations and non-trivial comprehensive improvements across all $15$ visual and language benchmarks.
\end{abstract}

\begin{IEEEkeywords}
visual languages, image analysis, data processing
\end{IEEEkeywords}

\section{Introduction}
\label{sec:intro}

Recent visual-aligned Large Multi-modal Models (LMMs) like MiniGPT4~\cite{zhu2023minigpt} and LLaVA~\cite{liu2023visual} have shown impressive capabilities in instruction-following and visual reasoning.
Most LMMs are built upon pre-trained Large Language Models (LLMs) and typically undergo a two-stage training process.
The pre-training stage establishes initial image-text alignment by training the LMM on large-scale image-text pairs, while the post-training stage—including visual instruction tuning and reinforcement-learning alignment\cite{rlhf-v,yu2024rlaifv}—better aligns LMMs with human intent.
During the visual instruction tuning stage, the pre-trained LLM within the LMM is unlocked to participate in training, facilitating a faster and more thorough alignment of modalities.
Consequently, visual instructions play a crucial role in shaping the capabilities of both the LMM and its underlying LLM, making the quality of these instructions critical for developing robust and powerful LMMs.

\begin{figure}[t!]
    \centering
    \includegraphics[width=0.95\linewidth]{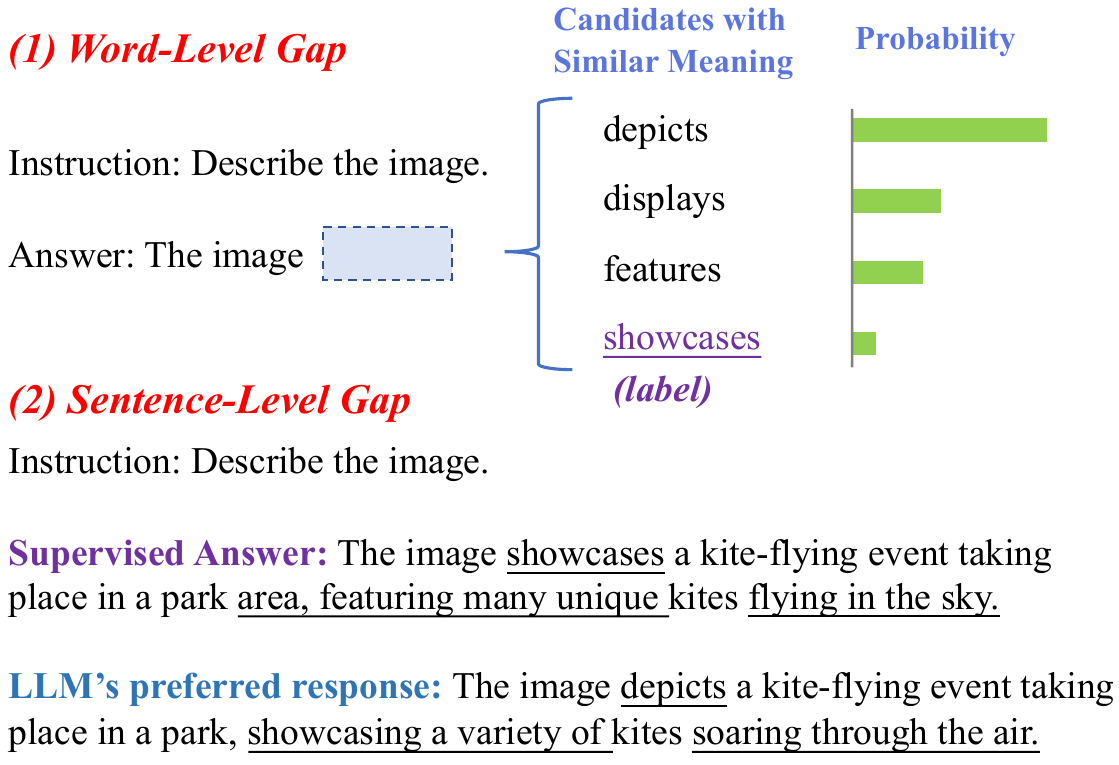}
    \caption{\textbf{The instance of word-level and sentence-level writing manner gap.}}
    \label{fig:instances}
\end{figure}

To enhance instructions, there are many efforts worked on building novel high-quality instruction datasets~\cite{li2023stablellava} or correcting factual errors in existing datasets~\cite{wang2023vigc,yu2023hallucidoctor}.
In this paper, different from them, we focus on assessing the instruction quality from a unique perspective called \textbf{Writing Manner}.
The writing manner refers to the specific habits of vocabulary selection, grammar usage, and sentence structuring used to express particular semantics.
We highlight a long-overlooked issue: there exists a severe \textbf{Writing Manner Gap} between the visual instructions and the inner pre-trained LLM, undermining the efficacy of LMMs.

In Figure~\ref{fig:instances}, we present the instance of writing manner gap at both the word and sentence levels for illustration.
Pre-trained LLM has its own unique writing style preferences, which are explicitly expressed in output probabilities of candidate tokens when generating new token.
The word-level gap arises when there are candidate words with similar meaning but higher probabilities than the labeled word.
Since LMMs are token-by-token probability predictors, the accumulation of word-level gap leads to the sentence-level gap, which is reflected in aspects of phrase, grammar, and sentence structure.
During the visual instruction tuning phase, the writing manner gap forces the LLM to change its original writing style, which may causes performance degradation or even catastrophic forgetting.
Therefore, to maintain the LLM performance and further build the robust LMM, it is essential to minimize the writing manner gap between the LLM and training instructions.

\begin{figure}[t]
    \centering
    \includegraphics[width=0.95\linewidth]{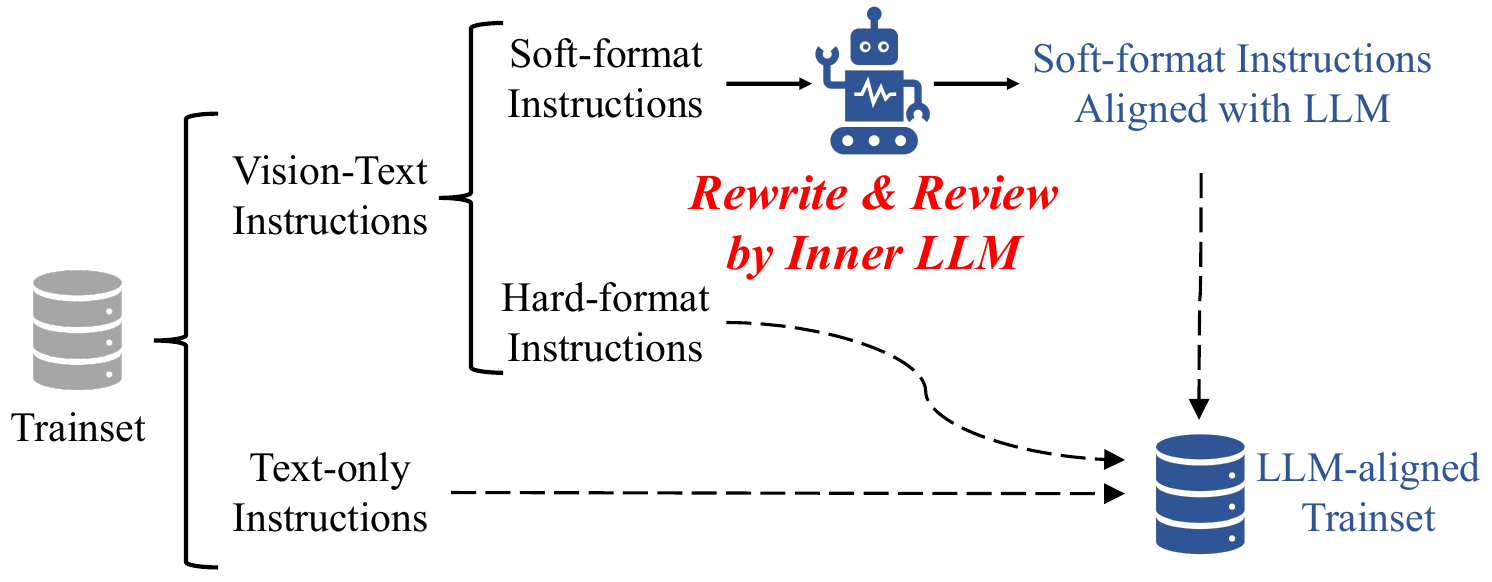}
    \caption{\textbf{The brief diagram of our LLM-aligned trainset construction. }}
    \label{fig:diagram}
    \vspace{-0.15in}
\end{figure}

In this paper, we propose a simple yet effective instruction processing approach to address this problem, as illustrated in Figure~\ref{fig:diagram}.
We leverage the pre-trained LLM within LMM to align the writing manner of soft-format visual instructions with that of the LLM itself, under the promise of keeping the original semantics of these instructions unchanged.
Soft-format visual instructions refer to open-ended question-answer pairs characterized by a high degree of freedom in textual expression, offering ample opportunities for adjustments and improvements.
Specifically, the answer part of soft-format visual instructions is first rewritten by the inner LLM to match its writing manner, and then reviewed by the inner LLM to ensure the alignment of writing manner is completed while preserving the original meaning.
If the revised answer is deemed unqualified during the review, the original answer is retained.
By combining these writing manner-aligned visual instructions with other remaining instructions, the proposed LLM-aligned trainset is created.

We adopt well-known LLaVA-1.5~\cite{liu2023improved} along with its trainset and QwenVL~\cite{bai2023qwenvl} as baseline models and trainset.
We design a human assessment procedure regard writing manner, and corresponding results demonstrate that our approach successfully realizes writing manner alignment.
By utilizing our LLM-aligned trainset, both LLaVA-7B and QwenVL achieve non-trivial comprehensive improvements across $15$ visual and language benchmarks.
Furthermore, careful cross-evaluation and ablation studies confirm that most improvements are brought by writing manner alignment rather than instruction revision.

Our contribution is three-folds:
\textbf{1) }To our knowledge, we are the first to identify the problem of writing manner gap between training instructions and pre-trained LLMs within LMMs.
\textbf{2) }Without introducing any external data or models, we leverage the inner LLM of LMM to reduce writing manner gap by rewriting and reviewing soft-format visual instructions.
\textbf{3) }Extensive experiments based on LLaVA-1.5 and QwenVL demonstrate the importance of reducing writing manner gap and the effectiveness of our approach.

\section{Related Work}

Various approaches have been proposed using traditional small models, such as detectors and OCR tools, to reduce factual errors and visual hallucinations or to create specialized visual instructions~\cite{zhang2023llavar,ye2023mplugdoc,liu2023mitigating}.
For example, HalluciDoctor~\cite{yu2023hallucidoctor} designed a cross-checking paradigm to cut down visual hallucinations, while LURE~\cite{zhou2023analyzing} evaluated underlying hallucinations based on co-occurrence, uncertainty and object position, and reconstructs less hallucinatory descriptions.
Another strategy related to ours leverages LLMs or LMMs to improve existing instructions.
In vision-language representation domain, LaCLIP~\cite{fan2023improving} and VeCLIP~\cite{lai2023scarcity} employed LLMs to rewrite or amalgamate image captions to enhance CLIP training.
Additionally, some methods~\cite{Zhao2023MLLMDataEngineAI,Du2023WhatMF} utilized powerful external LLMs or LMMs to clean or synthesize visual instructions.


In this paper, we focus on reducing the writing manner gap by rewriting visual instructions with the inner LLM of LMM.
Considering that our method ensures the original semantics remain unchanged, the proposed method complements other data augmentation and enhancement approaches.

\section{The Problem of Writing Manner Gap}



\subsection{Cause}
\label{subsec:causes pf gap}

The writing manner refers to the manifestation of writing style in terms of vocabulary, grammar, sentence structures, and other stylistic choices used to express particular semantics.

The writing manner of LLM is typically shaped during its post-training process, heavily influenced by the data and training methods used in this phase.
Different LLMs often exhibit distinct writing manners, which are mainly reflected in the following two aspects.
On one hand, when express particular meanings, LLMs perform differently in using vocabulary, grammar, sentence structure, and many other aspects.
On the other hand, given the same input context, responses generated by different LLMs may differ in semantic, length, writing level and so on.
A straightforward example is that some LLMs provide concise answers, while others are more verbose.

Therefore, when selecting a particular LLM to build the LMM, the inherent output characteristics of the LLM should not be overlooked.
However, existing strategies of multi-modal instruction trainset construction have not taken the above LLM properties into account.
Typically, the visual instruction datasets primarily originate from three sources: expert manual annotation; generation by advanced LLMs based on visual-related textual information; and the collection of outputs from LMMs.
Researchers directly employ the mixture of these data to directly train various kinds of LMMs, leading to an evident conflict between the writing manner of the training data and the inner LLM.


\subsection{Impact}
\label{subsec:impacts of gap}


During the visual instruction tuning stage, most LMMs facilitate the training of inner LLM to achieve faster and more thorough alignment between vision and language.
In this situation, the writing manner gap forces the LLM to change its original writing habits to match the writing style of the training data, which may cause severe capability degradation and even catastrophic forgetting.
Intuitively, the more pronounced the writing manner gap, the more the LLM is changed, leading to more severe capability degradation.

Furthermore, since the LLM within the LMM plays a central role in processing and integrating multi-modal information, it’s crucial to preserve its capabilities to build robust LMMs. 
When the writing manner gap causes LLM degradation, it also harms the LMM’s ability to generalize and respond accurately. 
This leads to more incorrect answers and visual hallucinations, especially when handling unfamiliar, open-domain visual tasks.

Therefore, the writing manner gap is detrimental to the performance of both the inner LLM and the LMM. 
Bridging the writing manner gap is an emergent and meaningful task.

\begin{figure*}[t!]
    \centering
    \includegraphics[width=0.98\linewidth]{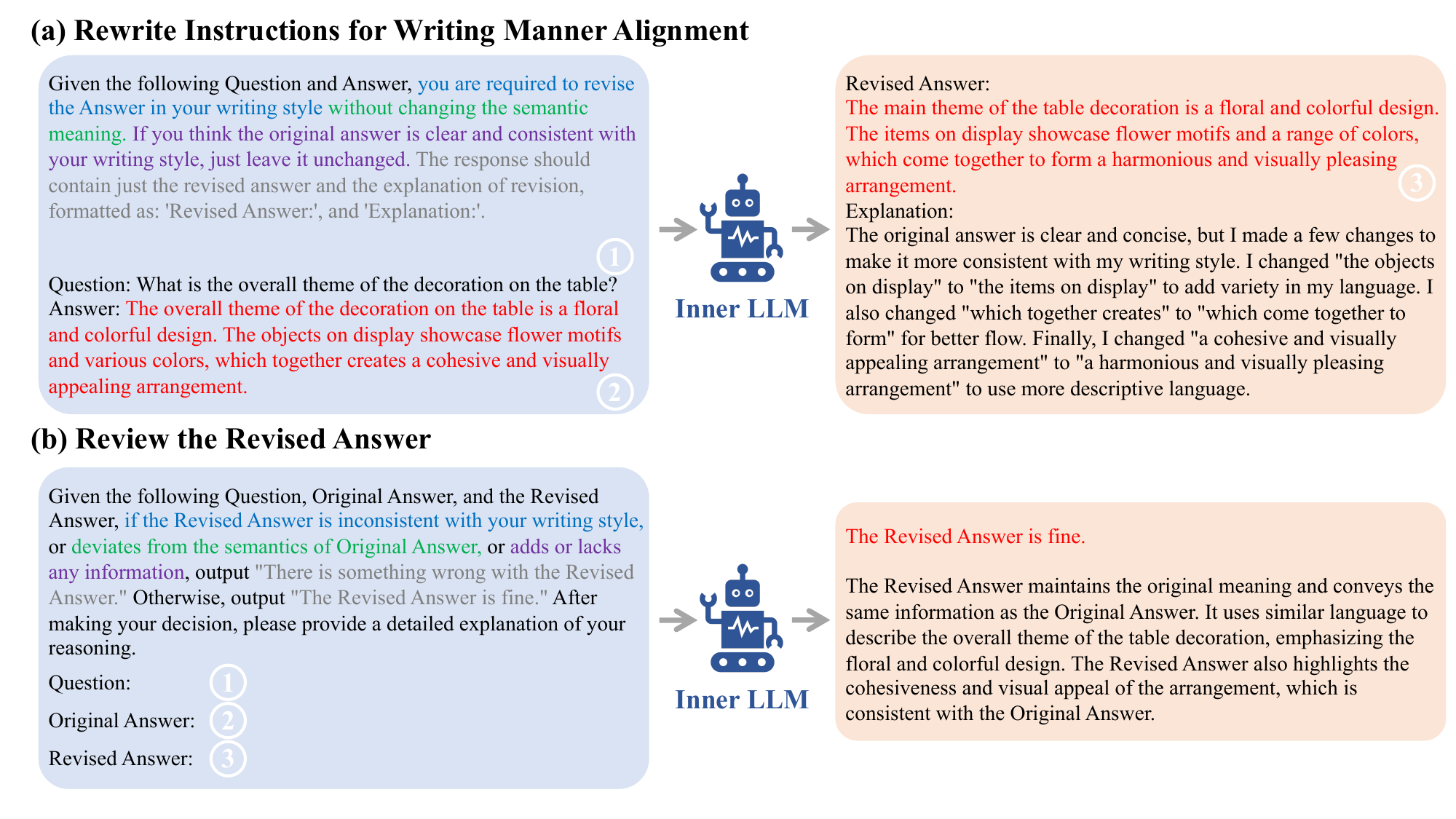}
    \caption{\textbf{One positive instance generation process for the illustration of writing manner alignment.} The important information are highlighted in non-black colors for clarity.}
    \vspace{-0.1in}
    \label{fig:alignExample}    
\end{figure*}

\section{Methodology}
\subsection{Overall Processing}

To narrow the writing manner gap, we propose directly utilizing the inner LLM to transfer the writing manner of soft-format visual instructions to align with that of the inner LLM itself under the promise of not changing original semantics.

This approach is feasible for two main reasons.
On one hand, thanks to excellent instruction-following and reasoning capabilities, LLM can intelligently answer questions posed by prompts that contain requirements and input information.
On the other hand, the responses generated by LLM naturally fall within the high probability regions of its output distribution space, which exactly meets with the purpose of reducing writing manner gap.

The specific instruction alignment process includes two stages: LLM rewriting and review.
The former realizes the writing manner transfer of original answers, while the latter is utilized for quality control, aimed at eliminating errors and anomalies in the modified answers.
Both of these processes operate at the level of single-round conversation, and do not require the input of visual features.
Figure~\ref{fig:alignExample} presents a detailed positive instance for illustration.


\subsection{Trainset Partition}

As shown in Figure~\ref{fig:diagram}, depending on the strictness of format requirements, the vision-text instructions in the trainset can be categorized into hard-format and soft-format instructions.

Hard-format instructions require answers written in a strict format, such as a single word or letter, a phrase, a coordinate, or a brief one-sentence description.
Many tasks, such as visual multiple-choice questions, true/false questions, OCR, and visual grounding, fall into this category.
Under the premise of not changing semantics, the room for modification in hard-format data is quite limited.
In contrast, soft-format instructions, such as open-ended questions and visual reasoning tasks, are tolerant of length, grammar, structure, as long as the content is logical and coherent.
Therefore, soft-format instructions have greater flexibility and are more amenable to be modified in writing manner.
Consequently, we perform writing manner alignment on soft-format instructions, and further mix them with hard-format instructions and text-only instructions to construct the LLM-aligned trainset.


    

\begin{table*}[!ht]
\centering
\resizebox{\linewidth}{!}{
\begin{tabular}{llc | ccccc| ccccccc }
\toprule
LMM & inner LLM & IT & VQA$^\text{v2}$ & GQA & VisWiz & SQA$^\text{I}$ & VQA$^\text{T}$ & POPE & MME & MMB & MMB$^\text{CN}$ & SEED$^I$ & LLaVA$^\text{W}$ & MM-Vet \\
\midrule
LLaVA & Vicuna-7B & Ori & 78.5 & 62.0 & 50.0 & 66.8 & 58.2 & 85.9 & 1510.7 & 64.3 & 58.3 & 66.2 & 63.4 & 30.5 \\
 LLaVA & Vicuna-7B & Ours & \textbf{79.1} & \textbf{62.9} & \textbf{51.3} & \textbf{71.3} & \textbf{58.8} & \textbf{87.2} & \textbf{1513.0} & \textbf{66.6} & \textbf{59.7} & \textbf{67.0} & \textbf{67.5} & \textbf{31.9} \\
 
\midrule

LLaVA & Vicuna-13B & Ori & \textbf{80.0} & 63.3 & 53.6 & \textbf{71.6} & \textbf{61.3} & 85.9 & 1531.3 & \textbf{67.7} & \textbf{63.6} & 68.2 & 70.7 & 35.4 \\
 LLaVA & Vicuna-13B & Ours & \textbf{80.0} & \textbf{63.6} & \textbf{54.3} & \textbf{71.6} & \textbf{61.3} & \textbf{87.4} & \textbf{1569.7} & 67.3 & 63.0 & \textbf{68.5} & \textbf{72.9} & \textbf{36.6} \\

\midrule

QwenVL & Qwen-7B & Ori & 81.2 & 63.0 & 50.8 & 71.5 & 62.6 & 87.1 & 1576.8 & 71.8 & 64.6 & 68.4 & 70.5 & 41.7 \\
 QwenVL & Qwen-7B & Ours & \textbf{81.4} & \textbf{63.1} & \textbf{51.0} & \textbf{71.6} & \textbf{62.9} & \textbf{87.2} & \textbf{1589.3} & \textbf{72.0} & \textbf{65.0} & \textbf{68.9} & \textbf{72.3} & \textbf{44.3} \\

\midrule
\midrule
LLaVA & Vicuna-7B$^*$ & Ori & 69.5 & 47.2 & 40.3 & 57.3 & 39.7 & 84.1 & 1104.1 & 45.5 & 32.6 & 50.2 & 58.8 & 28.9 \\
 LLaVA & Vicuna-7B$^*$ & Ours & \textbf{69.7} & \textbf{47.6} & \textbf{43.3} & \textbf{58.1} & \textbf{39.8} & \textbf{85.2} & \textbf{1161.1} & \textbf{46.8} & \textbf{34.7} & \textbf{50.6} & \textbf{59.4} & \textbf{29.8} \\
 
\bottomrule
\end{tabular}
}
\caption{\textbf{Performance comparisons of baseline models on 12 visual benchmarks.} 
``IT'' indicates the trainset used in instruction tuning stage, where ``Ori'' refers to the original trainset of LLaVA-1.5 and ``Ours'' means the LLM-aligned trainset proposed in this paper.
The $*$ represents the inner LLM is frozen during the fine-tuning.
}
\vspace{-2mm}
\label{tab:comparisons}
\end{table*}

\subsection{Align Instructions with LLM}


\noindent\textbf{LLM Rewriting Stage.}
The input text in LLM rewriting stage consists of three parts: the requirement for writing manner alignment, the question, and the answer.
Note that because the question in conversation represents the user's intent, it does not need to be modified.
Providing the question to LLM enables it to better understand the context of the conversation.
As shown in Figure~\ref{fig:alignExample}, the task requirement part should include four key points: 
1) Request the rewriting of the original answer to align with the writing manner of the LLM; 
2) Emphasize that such rewriting must not change the semantic meaning; 
3) Specify that if the original answer already conforms to the LLM's writing manner, no modification is necessary;
4) Specify the format of responses.
Afterwards, the post-processing operation is designed to separate the revised answer from the LLM response.
The status yielded by post-processing step indicates whether the desired answer has been obtained.

\noindent\textbf{LLM Review Stage.} 
Due to the randomness of LLM's output and the risk of rewriting failure, we utilize the LLM to review the modified answers for quality assurance.
To ensure the certainty of the review, sampling operations are disabled during LLM inference.
The review prompt includes four parts: the review requirement, the question, the original answer, and the revised answer.
A positive review judgement must meet two criteria: 
1) The revised answer does not change the semantics of the original answer, nor does it omit or add content;
2) The revised answer is well-aligned with the writing manner of the LLM.
Notably, when making a decision, we require the LLM to output specific judgement sentences, rather than just a word like ``Yes/No'', to improve the accuracy of review determinations.
Only when the revised answer passes the review is it used to replace the original answer.

\section{Experiments}

\subsection{Setting}

\noindent\textbf{Baseline Models and Dataset.} In this paper, we utilize the well-known LLaVA-1.5 and QwenVL as the baseline model.
LLaVA-1.5 employs the Vicuna-1.5 as the inner LLM, offering two versions of 7B and 13B parameters, while the QwenVL deploys the Qwen-7B as the inner LLM.

Considering LLaVA-1.5's exceptional performance and its recognition within the industry, we uniformly adopt the LLaVA-1.5's trainset as the visual instruction trainset for both LLaVA-1.5 and QwenVL pre-trained models.
The writing manner of soft-format visual instructions in trainset are aligned with inner LLMs for quality enhancement.

LLaVA's training dataset is a mixture of public available academic task-oriented data~\cite{marino2019ok,schwenk2022okvqa,mishra2019ocr,sidorov2020textcaps,krishna2017visual,kazemzadeh2014referitgame,sharegpt}.
According to answer format, we split the visual instructions into five types, which are visual conversations, one word/phrase VQA, choice questions, short captions, and groundings.
Visual conversations are open-ended, belong to the soft-format category, while the latter four types are restricted or brief, falling into hard-format category.
Therefore, the data eligible for adjustment is visual conversation data, totaling 158K, which approximately constitutes a quarter of overall visual instructions.

\noindent\textbf{Implementation Details.} We implement the visual instruction alignment and model training using $8 \times$ A800s.
To increase the throughput and accelerate inference speed, we utilize the vLLM framework~\cite{kwon2023efficient} to load and run LLMs.
There are a total of $361$K rounds of conversations for soft-format visual instructions. 
By combining original text-only instructions, hard-format visual instructions and LLM-aligned soft-format visual instructions, the novel LLM-aligned trainset is formed.
To ensure fairness, the order of training instructions is consistent with LLaVA-1.5, and the training hyper-parameters are same with official settings of LLaVA-1.5 and QwenVL.

\subsection{Writing Manner Alignment Assessment}

We conduct human evaluation to verify whether the inner LLM can reliably achieve writing manner alignment during the rewriting stage in the following three steps:
i) Selecting samples to represent the respect writing manner of inner LLM and original dataset: For the inner LLM, we randomly select 20 responses to questions from the Alpaca-eval benchmark; for the original instruction dataset, we randomly select 20 soft-format visual instructions.
ii) Selecting samples to be evaluated: We randomly selected 100 soft-format visual instructions modified by inner LLM as evaluation samples.
iii) Human evaluation: We invite four human experts to determine which writing manner the samples are more similar to. The options include: ``inner LLM'', ``original dataset'' and ``none of both''.

According to Table~\ref{tab:human_evaluation}, both inner LLMs participating in this assessment achieved a success rate of over 90\% in writing manner alignment task, indicating that utilizing inner LLM for writing manner alignment is highly reliable. 
Additionally, Qwen-7B outperforms Vicuna-7B, possibly due to the more significant writing manner gap between Qwen-7B and original visual instruction dataset.

\begin{table*}[t]
  \centering
  \begin{minipage}[t]{0.35\textwidth}
    \centering
    \scalebox{0.85}{
    \tabcolsep2pt
    \renewcommand{\arraystretch}{1.5}
    \begin{tabular}{ccccc}
    \toprule
    \multirow{2}[2]{*}{LMM} & \multirow{2}[2]{*}{Inner LLM} & \multicolumn{3}{c}{Voting Options} \\
          &       & inner LLM & original dataset  & none of both \\
    \midrule
    LLaVA & Vicuna-7B & 90.5\% & 8.5\%   & 1.0\% \\
    QwenVL & Qwen-7B & 94.5\% & 5.5\% & 0\% \\
    \bottomrule
    \end{tabular}%
    }
    \caption{The average voting results of manual writing manner assessment.}
    \label{tab:human_evaluation}%
  \end{minipage}%
  \hfill
  \begin{minipage}[t]{0.3\textwidth}
    \centering
    \scalebox{0.8}{
    \begin{tabular}{cc|c|cc}
    \toprule
    \multirow{2}[2]{*}{Model} & \multirow{2}[2]{*}{IT} & \multirow{2}[2]{*}{Pope} & \multicolumn{2}{c}{HallusionBench} \\
          &       &       & Figure Acc & Question Acc \\
    \midrule
    LLaVA-7B & Ori   & 85.9  & 14.16 & 44.82 \\
    LLaVA-7B & Ours  & \textbf{87.2}  & \textbf{16.19} & \textbf{46.32} \\
    \midrule
    QwenVL & Ori   & 87.1  & 16.47 & 42.69 \\
    QwenVL & Ours  & \textbf{87.2}  & \textbf{19.08} & \textbf{43.14} \\
    \bottomrule
    \end{tabular}%
    }
    \caption{Visual and textual hallucination evaluation.}
    \label{tab:hallucination}%
  \end{minipage}%
  \hfill
  \begin{minipage}[t]{0.3\textwidth}
    \centering
    \scalebox{0.8}{
    \renewcommand{\arraystretch}{1.2}
    \begin{tabular}{cc|cc}
    \toprule
    Model & IT    & MTBench & Alpaca-eval \\
    \midrule
    LLaVA-7B & Ori   & 5.98  & 5.19 \\
    LLaVA-7B & Ours  & \textbf{6.04}  &\textbf{ 5.28} \\
    \midrule
    QwenVL & Ori   & 4.89  & 2.99 \\
    QwenVL & Ours  & \textbf{5.01}  & \textbf{3.16} \\
    \bottomrule
    \end{tabular}%
    }
    \caption{Performance comparisons on LLM evaluation benchmarks.}
    \label{tab:llm_eval}%
  \end{minipage}
\end{table*}

\begin{table*}[!t]
  \centering
    \tabcolsep2pt
    \resizebox{0.99\textwidth}{!}{ 
    \begin{tabular}{l ccc|ccccc|ccccccc}
    \toprule
    Model & w/o Soft & Rewrite & Review & VQA$^\text{v2}$ & GQA   & VisWiz & SQA$^\text{I}$   & VQA$^\text{T}$  & POPE  & MME   & MMB   & MMB$^{\text{CN}}$ & SEED$^I$  & LLaVA$^\text{W}$ & MM-Vet \\
    \midrule
    \multirow{4}[2]{*}{LLaVA-7B} &       &    &   & 78.5  & 62.0  & 50.0  & 66.8  & 58.2  & 85.9  & 1510.7 & 64.3  & 58.3  & 60.1  & 63.4  & 30.5 \\
    &$\surd$       &    &   &  78.8 &  62.2 & 48.4  & 68.1  & 57.5  & 86.6  & 1502.6 & 66.8  & 58.8  &   66.1 & 50.0  & 29.0 \\
         & & $\surd$     &       &   \textbf{79.1}   &   62.8   &   50.7    & 69.6  & 58.6  & 87.1  & 1488.5 & \textbf{67.0}  & \textbf{60.4}  &   66.2   &   \textbf{68.6}   &  \textbf{33.1} \\
         & & $\surd$     & $\surd$     & \textbf{79.1}  & \textbf{62.9}  & \textbf{51.3}  & \textbf{71.3}  & \textbf{58.8}  & \textbf{87.2}  & \textbf{1513.0} & 66.6  & 59.7  & \textbf{67.0}  & 67.5  & 31.9 \\

    \bottomrule
    \end{tabular}%
    }
  \caption{\textbf{The ablation study of soft-format visual instructions, LLM rewrite and review stage.}}
  \label{tab:ablation}%
  \vspace{-0.10in}
\end{table*}%

\subsection{Visual Performance Comparisons}

\noindent\textbf{Comparison with Baseline.} 
The quantitative comparison results are shown in Table~\ref{tab:comparisons}.
By training with our LLM-aligned trainset, LLaVA-7B and QwenVL significantly improve the performance on all benchmarks, while LLaVA-13B achieves performance enhancements in $10$ out of $12$ benchmarks.

The soft-format training instructions directly impact model performance in open-ended question-answering scenarios.
The improvements observed in two baseline models on LLaVA$^\text{W}$ and MM-Vet benchmarks demonstrate the efficacy of our approach in enhancing data quality, which positively influences the training process.
Furthermore, improvements on academic benchmarks indicate a reduction in domain conflicts between different instruction sources in trainset, and might also be attributed to the strengthened maintenance effect of our LLM-aligned trainset on the capabilities of LLM, thereby bolstering the comprehension abilities of LMM.

Moreover, we also investigate the impact of LLM-aligned trainset to the LMM with frozen inner LLM.
According to the last two lines in Table~\ref{tab:comparisons}, LLaVA-7B achieves comprehensive improvements once again, which indicates that LLaVA-7B performs better convergence extent to LLM-aligned trainset than original trainset.

\noindent\textbf{Hallucination Evaluation.}
Hallucinations seriously impair the usability of LMMs.
To investigate the impact of the proposed LLM-aligned instruction set on model hallucinations, we conduct hallucination assessments using POPE and HallusionBench~\cite{guan2023hallusionbench}, with the corresponding results presented in Table~\ref{tab:hallucination}.
The comparisons indicate that our method effectively enhances the LMM's accuracy in both visual and textual scenarios.
Recalling the analysis in Subsection~\ref{subsec:impacts of gap}, our method successfully reduces the writing manner gap, thereby mitigating the disturbances to the inner LLM during the visual instruction tuning stage and improving the LMM performance.

\subsection{Textual Performance Comparisons}

\noindent\textbf{Comparison with Baseline.} 
We evaluate the performance of LMMs in textual scenarios by using MTBench~\cite{zheng2023judging} and Alpaca-eval~\cite{alpaca_eval}.
There two benchmarks utilize GPT-4 to score or rank model answers compared with reference answers.
Table~\ref{tab:llm_eval} displays the scores of LMMs trained with different instruction sets on MTBench (where the mean score of two assessments is taken here to mitigate the randomness of GPT-4 scoring), as well as win rates on Alpaca-eval.
On both benchmarks, LLM-aligned trainset bring improvements to all baseline models compared with original instructions, demonstrating that our approach effectively alleviates the LLM degradation caused by soft-format visual instructions.

\subsection{Ablation Study}

\noindent\textbf{The Influence of Soft-Format Instructions.} We deploy the combination of text-only and hard-format instructions for fine-tuning to explore the influence of soft-format visual instructions.
We keep the same training steps to ensure comparison fairness.
According to results in Line $2$ of Table~\ref{tab:ablation}, without soft-format training instructions, the model achieves comparable or even better performance in VQA benchmarks, but drops a lot in open-ended benchmarks.
The result indicates the soft-format visual instructions primarily contribute to enhancing the model's performance in open-ended scenarios.
Moreover, there are domain conflicts between the soft-format and hard-format instructions, lies in aspects such as task type, correctness, and writing manner.
Minimizing the domain conflict is beneficial for improving the model's general capabilities.

\noindent\textbf{The Effectiveness of Rewrite \& Review.} 
Table~\ref{tab:ablation} presents the ablation results of LLM rewrite and review stages.
With the rewritten instructions, model performs better on all benchmarks except MME.
The LLM review stage further filtered out unqualified rewritten instructions, leading to better performance in VQA tasks.
There are slight declines in open-ended visual tasks compared to with only rewriting stage, which may attributed to the potential conflicts caused by directly replacing unqualified revised answers with original answers.

\begin{table}[!t]
  \centering
    \resizebox{0.49\textwidth}{!}{
    \begin{tabular}{lc|cccc}
    \toprule
    LMM   & IT   & GQA   & VQA$^{T}$ & MMB   & LLaVA$^W$ \\
    \midrule
    LLaVA-7B & Original &  62.0  & 58.2  & 64.3  & 63.4 \\
    LLaVA-7B & Self-aligned &  62.9  & 58.8  & 66.6  & 67.5 \\
    LLaVA-7B & Cross-aligned &  62.4  & 57.9  & 64.4  & 63.8 \\
    \midrule
    QwenVL & Original  & 63.0  & 62.6  & 71.6  & 70.5 \\
    QwenVL & Self-aligned  & 63.1  & 62.9  & 72.0  & 72.3 \\
    QwenVL & Cross-aligned  & 61.8  & 61.9  & 71.3  & 71.0 \\
    \bottomrule
\end{tabular}%
    }
  \caption{\textbf{Cross-evaluation results for sanity check.} The ``Cross-aligned'' means the trainset is aligned by the other LMM, either LLaVA-7B or QwenVL.}
  \label{tab:crosseval}%
  \vspace{-0.10in}
\end{table}%

\noindent\textbf{Sanity Check by Cross-Evaluation.}
We design a cross-evaluation experiment to determine whether the improvements are primarily due to bridging the writing manner gap rather than enhancements from LLM revision.
Specifically, we train LLaVA-7B using the Qwen-7B-aligned trainset and Qwen-VL using the Vicuna-7B-aligned trainset, with the results shown in Table~\ref{tab:crosseval}.
In this cross-evaluation setup, there is noticeable writing manner gap between trainsets and models.
Given that both models are improved by their respective aligned trainsets, if the cross-evaluation shows better performance, we can infer that the LLM revision is the key factor. 
If not, it indicates that reducing the writing manner gap is crucial.
As seen in Table~\ref{tab:crosseval}, compared to using the original trainset, the LLaVA-7B in the cross-evaluation setting shows slight fluctuations, while Qwen-VL with cross-aligned trainset exhibits significant performance drop on GQA and VQA$^T$ benchmarks.
This result strongly demonstrates the importance and effectiveness of reducing writing manner gap.

\section{Conclusion}

In this paper, we highlight the issue of the writing manner gap between the visual instruction trainset and the inner LLM of LMM.
The writing manner gap severely hinder the development of robust LMMs.
Without introducing any external data or models, we leverage the inner LLM to bridge writing manner gap.
Experimental results validate the effectiveness of our motivation and methodology.

\bibliographystyle{IEEEbib}
\bibliography{custom}

\clearpage
\appendix
\label{sec:appendix}
\subsection{Setting}

\begin{table}[!t]
  \centering
  \scalebox{0.95}{
    \tabcolsep3pt
    \resizebox{\linewidth}{!}{
    \begin{tabular}{c|c|ccc}
    \toprule
    LMM   & LLM   & Rewrite & Review & Instruction Tuning \\
    \midrule
    LLaVA & Vicuna-7B & $\sim$ 10h & $\sim$ 10h  & $\sim$ 10h \\
    LLaVA & Vicuna-13B & $\sim$ 15h & $\sim$ 15h  & $\sim$ 20h \\
    QwenVL & Qwen-7B & $\sim$ 5h & $\sim$ 5h & $\sim$ 22h \\
    \bottomrule
    \end{tabular}%
    }
  }
    \caption{\textbf{Time overheads} for soft-format visual instruction writing manner alignment and visual instruction tuning by using $8\times$ A800s.}
  \label{tab:timecost}%
\end{table}%

\subsubsection{Evaluation Benchmarks} 
\label{subsec:benchmark}
By utilizing LLM to transfer writing manner of visual instructions, our approach involves a trade-off between minimizing the writing manner gap and introducing noise.
To validate that our method prioritizes the former and that the impact of noise is limited, we evaluated models on $12$ benchmarks for thorough assessment.

VQA$^{v2}$~\cite{goyal2017making}, GQA~\cite{hudson2019gqa}, VisWiz~\cite{gurari2018vizwiz}, SQA$^I$~\cite{lu2022learn}, VQA$^T$~\cite{singh2019towards} are academic benchmarks in the realm of traditional Visual Question Answering (VQA) tasks.
POPE~\cite{li2023evaluating} is a polling-based query benchmark for evaluating the vision hallucination.
The MME~\cite{fu2023mme} benchmark evaluates LMM's perception and cognition capabilities through a series of carefully crafted questions across $14$ sub-tasks.
MMBench and MMBench-CN~\cite{liu2023mmbench} manually design questions in English and Chinese to evaluate model's vision reasoning ability.
SEED~\cite{li2023seed} benchmark is constructed with the assistance of GPT4, covering scenes in images and videos.
Due to the absence of some video sources, we employ SEED's image part for evaluation.
LLaVA (in the wild)~\cite{liu2023visual} and MM-Vet~\cite{yu2023mm} are open-ended benchmarks, which use GPT4 for LMM capability assessment.

\subsubsection{Hyperparameters}
In Table~\ref{tab:hyperparameters}, we show the generation hyperparameters in LLM rewriting and review stage.
During the instruction tuning stage, we use the same set of hyper-parameters as the original LLaVA-1.5~\cite{liu2023improved} and QwenVL~\cite{bai2023qwenvl}.


\begin{table}[t!]
  \centering
    \resizebox{\linewidth}{!}{
    \begin{tabular}{c|c|cccc}
    \toprule
    LLM & Stage & Temperature & top\_p & top\_k & max\_length \\
    \midrule
    Vicuna & rewriting & 0.4   & 0.6   & 5     & 2048 \\
    Qwen & rewriting  & 0.2     & 0.6     & 5     & 2048 \\
    \bottomrule
    \end{tabular}%
    }
  \caption{Generation configurations of writing manner alignment.}
  \label{tab:hyperparameters}%
\end{table}%

\begin{table}[t]
  \centering
    \tabcolsep3pt
    \resizebox{0.95\linewidth}{!}{
    \begin{tabular}{c|c|c|cc}
    \toprule
    Model & LLM & Total QA & Failures & Unqualified Samples \\
    \midrule
    LLaVA & Vicuna-7B & \multirow{3}[2]{*}{361K} & 0.4K (0.11\%)  & 2K (0.55\%) \\
    LLaVA &  Vicuna-13B  &   & 0.7K (0.19\%) & 3.5K (0.97\%) \\
    QwenVL & Qwen-7B &  & 0.3K (0.08\%) & 0.8K (0.22\%) \\
    \bottomrule
    \end{tabular}%
    }
  \caption{The quantity of failure cases in rewriting stage and unqualified samples in review stage.}
  \label{tab:numbers}%
\end{table}%

\subsubsection{Post-process Step}
\label{sec:post_process}
\noindent\textbf{Procedure.} The objective of post-processing is to separate the desired answers from the responses of LLM and to filter out apparent errors.
The post-process step in LLM rewriting step contains two aspects.
Firstly, based on the prompt depicted in Fig.3, the response of LLM is expected to contain two segments, starting with ``Revised Answer:'' and ``Explanations: ''.
The portion between these two keywords is the desired modified answer.
If these keywords are absent, the attempt is considered a rewrite failure.
Secondly, we detect the presence of certain sensitive words that indicate obvious errors in the modified answer.
If these sensitive words are found, this rewrite is deemed a failure.
The sensitive words include ``revised answer'', ``original answer'', ``revision'', ``semantic meaning'', and ``Question''.
In cases of rewrite failure, the original answers are reserved.


\noindent\textbf{Statistics.}
Table~\ref{tab:numbers} presents numbers of failures in the rewriting stage and unqualified samples from the review stage.
The statistics reveal a extremely high success rate for data rewriting, with a tiny proportion of revised answers (less than $1\%$) be ing deemed unqualified during review.
Upon examining the quality of the revised answers, we found that Vicuna13B tend to over-elaborate, producing redundant words or sentences that were difficult to segment.
As reflected in the Table~\ref{tab:numbers}, compared to Vicuna-7B and Qwen-7B, Vicuna-13B has a higher error probability, leads to relatively lower improvement of LLaVA.
These findings suggest that our method places high demands on the instruction-following ability of LLMs.

\subsubsection{Details of Trainset}

We present specific compositions and quantities of trainset for instruction tuning in Table~\ref{tab:trainset}.
According to the answer format, we could split the visual instructions into five types, which are visual conversations, one word/phrase VQA, choice questions, short captions, and groundings.
In this work, visual conversations are aligned by inner LLM.

\subsection{More Experiments}

\subsubsection{Quantitative Measurement of Writing Manner Gap}

\textbf{Perplexity-Based Indicator.} 
To quantitatively measure the writing manner gap between the visual instruction set and the inner LLM, we propose a PPL-based indicator.
To begin with, given a tokenized sequence $X = (x_0, x_1, ..., x_t)$, the PPL of $X$ is computed as
\begin{equation}
    PPL(X) = exp\{ - \frac{1}{t} \sum_i^t log p_{\theta}(x_i | x_{<i}) \},
\end{equation}
where $log p_{\theta}(x_i | x_{<i})$ is the log-likelihood of the i-th token conditioned on the preceding tokens $x_{<i}$ according to model.
Intuitively, the PPL evaluates the model's ability to predict uniformly among the set of specified tokens in a corpus.

Assuming there is a pre-trained LMM $M$ and a visual instruction set $S$ which is divided into training set $S_t$ and evaluation set $S_e$, the proposed metric is obtained in two steps.
We first freeze the inner LLM of $M$ and train $M$ on $S_t$ till convergence to get $M'$, and then calculate the PPL score of the $M'$ only on the answer part of $S_e$.

\twocolumn[{%
\renewcommand\twocolumn[1][]{#1}%
\begin{center}
  \scalebox{0.84}{
    \tabcolsep2pt
    \begin{tabular}{c|ccc|cccc|c|c|cc|c}
    \toprule
    \multirow{2}[4]{*}{Type} & \multicolumn{3}{c|}{Soft-Format visual instructions} & \multicolumn{8}{c|}{Hard-Format visual instructions}          & Text-Only \\
\cmidrule{2-13}          & \multicolumn{3}{c|}{Visual Conversations} & \multicolumn{4}{c|}{One word or phrase VQA} & Choice & Short Caption & \multicolumn{2}{c|}{Grounding} & Conversation \\
    \midrule
    Data  & LLaVA Conv & LLaVA Detail & LLaVA Complex & VQAv2 & GQA   & OKVQA & OCRVQA & A-OKVQA & TextCaps & RefCOCO & VG    & ShareGPT \\
    Size  & 58K   & 23K   & 77K   & 83k   & 72K   & 9K    & 80K   & 50K   & 22K   & 30K   & 86K   & 40K \\
    \bottomrule
    \end{tabular}%
    }
\captionof{table}{\textbf{Data compositions of LLaVA-1.5 trainset.}}
\label{tab:trainset}
\end{center}%
}]

Why can this indicator represent the writing manner gap?
When the inner LLM is frozen, its inherent writing manner remains unchanged during training.
In this way, the LMM controls the subsequent output of the inner LLM sorely by adjusting visual prompts.
When the LMM converges on the training set $S_t$, it indicates that the model has aligned as closely as possible with the content and style of the training set.
At this point, the PPL measures how well the inner LLM accepts the style of the dataset.
Therefore, for a specific LMM, the smaller the PPL score brought by dataset, the closer that dataset's writing manner is to that of the inner LLM.

\begin{table}[!t]
  \centering
\resizebox{0.65\linewidth}{!}{
    \begin{tabular}{c|cc}
    \toprule
    \multirow{2}[4]{*}{Model} & \multicolumn{2}{c}{Soft-format Instructions} \\
\cmidrule{2-3}          & Original & LLM-aligned \\
    \midrule
    LLaVA-7B & 3.413 & \bf3.298 \\
    QwenVL & 4.208 & \bf3.932 \\
    \bottomrule
    \end{tabular}%
    }
  \caption{\textbf{PPL indicator of writing manner gap}.}
  \label{tab:ppl_indicator}%

\end{table}%

\noindent\textbf{Results and Analysis.} We utilize only the original and LLM-aligned soft-format visual instructions to conduct the aforementioned evaluation to the LLaVA-7B and QwenVL, where the last $3,000$ data entries serve as the $S_e$, and the remaining instructions constitutes the $S_t$.

According to the results in Table~\ref{tab:ppl_indicator}, both LLaVA-7B and QwenVL achieve lower PPL scores on the LLM-aligned instructions compared with the original instructions, indicating that our approach effectively reduces the writing manner gap.
Additionally, there is an interesting contrast that QwenVL exhibits higher PPL scores compared to LLaVA-7B.
This is because the trainset of LLaVA-7B's inner LLM Vicuna and the current soft-format visual instruction set both originate from ChatGPT, whereas QwenVL's inner LLM Qwen-7B performs a significant writing manner difference from soft-format visual instructions.

\begin{table}[t!]
  \centering
    \scalebox{0.8}{
    \begin{tabular}{p{13mm}|p{70mm}}
    \toprule
    Prompt Number & Content \\
    \midrule
    No.1     & Given the following Question and Answer, you are required to revise the Answer in your writing style without changing the semantic meaning. If you think the original answer is clear and consistent with your writing style, just leave it unchanged. The response should contain just the revised answer and the explanation of revision, formatted as: 'Revised Answer:', and 'Explanation:'. \\
    \midrule
    No.2     & Giving the following Question and Answer, you are required to accurately revise the answer to align with your writing style. Do not change its meaning. If you think the answer is clear, do not change it. The response should contain both the revised answer and corresponding explanation, formatted as 'Revised Answer:', and 'Explanation:'. \\
    \midrule
    No.3     & Giving the following Question and Answer, you are required to accurately revise the answer to align with your writing style. Do not change its meaning. If you think the answer is clear and consistent with your writing style, do not change it. The response should contain both the revised answer and corresponding explanation, formatted as 'Revised Answer:', and 'Explanation:'. \\
    \bottomrule
    \end{tabular}%
    }
  \caption{Rewriting prompts in stability validation.}
  \label{tab:prompt_rewrite}%
  \vspace{-0.05in}
\end{table}%

\begin{table}[t]
  \centering
    \scalebox{1.1}{
    \tabcolsep2pt
    \begin{tabular}{c|c|cccc}
    \toprule
    Model & Prompt Num & SQA   & POPE  & MMB   & LLaVA$^\text{W}$ \\
    \midrule
    \multirow{4}[2]{*}{LLaVA-1.5 7B} & -     & 66.8  & 85.9  & 64.3  & 63.4 \\
          & No.1    & 71.3  & 87.2  & 66.6  & 67.5 \\
          & No.2     & 68.7  & 86.9  & 67.3  & 69.8 \\
          & No.3     & 68.7  & 86.7  & 66.3  & 67.4 \\
    \bottomrule
    \end{tabular}%
    }
  \caption{The stability validation results of using rewriting prompts in same meaning but different expressions.}
  \label{tab:ablation_prompts}%
  \vspace{-0.14in}
\end{table}%

\subsubsection{The Stability Validation} 
\label{subsec:ablation_prompts}
Consider that outputs of LLMs have randomness and are heavily affected by prompts, we employ three different prompts with same meaning but varied wording in the rewriting stage to assess the stability of the proposed method.
The prompts are shown in Table~\ref{tab:prompt_rewrite} in Appendix, with corresponding results shown in Table~\ref{tab:ablation_prompts}.
The evaluation results on four representative benchmarks indicate that the LLM-aligned trainset consistently improves LLaVA's performance, although the extent of the improvement exhibits some variability.

\subsubsection{Comparison with Other Revision Strategies.}
To further validate the effectiveness of bridging the writing manner gap, we compare the default setting with two different rewriting prompts.

The first strategy specifies a particular writing style of ``plain English'' by replacing the ``your writing style'' in default prompt with ``plain English as you explain it to your children''.
In Table~\ref{tab:compare_with_revision}, we see that aligning the trainset’s style to “plain English” results in larger PPL score than using the original trainset, from 3.413 to 3.465, which indicates that this style significantly differs from the default writing style of inner LLM.
As for downstream evaluations, this revision method leads to poor performance on LLaVA$^W$ benchmark.

The second strategy lets LLM just revise the answer without any writing manner alignment requirements for ablation.
In this setting, we remove ``in your writing style'' and ``and consistent with your writing style'' in default prompt.
As shown in Table~\ref{tab:compare_with_revision}, the inner LLM naturally generates responses similar to its default writing manner when there is no writing style constraint, indicated by PPL score drops from 3.413 to 3.395.
However, this PPL decrease is not as significant as using the proposed rewriting prompt, confirms the necessity of adding writing manner alignment constraints for better reducing writing manner gap.
The downstream evaluations show that this strategy enhances the model performance on most downstream tasks, except for VQA$^T$ benchmark.
By comparison, the overall improvement brought by these two competing strategies is far more lower than that of the proposed method, which strongly validates the importance of writing manner alignment.

\begin{table}[!t]
  \centering
  \renewcommand{\arraystretch}{1.2}
    \resizebox{0.49\textwidth}{!}{
    \begin{tabular}{lc|c|cccc}
    \toprule
    LMM   & IT    & PPL$\downarrow$   & GQA   & VQA$^{T}$ & MMB   & LLaVA$^W$ \\
    \midrule
    LLaVA-7B & Original & 3.413 & 62.0  & 58.2  & 64.3  & 63.4 \\
    LLaVA-7B & ``Plain English'' Style & 3.465 & 62.2  & 58.3  & 64.6  & 62.5 \\
    LLaVA-7B & Revision \& No Align & 3.395 & 62.4  & 57.9  & 66.4  & 66.2 \\
    LLaVA-7B & Self-aligned (Ours) & \textbf{3.298} & \textbf{62.9}  & \textbf{58.8}  & \textbf{66.6}  & \textbf{67.5} \\
    \bottomrule
\end{tabular}%
    }
  \caption{\textbf{Comparison with two other revision strategies:}
  1) Specific writing style of ``plain English''; 2) Just revision with no writing manner alignment requirement.}
  \label{tab:compare_with_revision}%
  \vspace{-0.10in}
\end{table}%

\subsection{Discussion on Implementation Details}

\subsubsection{Why the generated explanations in LLM rewriting stage are not used afterward?}

We have attempted to instruct LLMs to directly output the rewritten answers without any additional information, but their instruction-following abilities are not strong enough.
LLMs always append some extra explanations after outputting the revised answer, which hinders the subsequent extraction of the desired answers from the LLMs' responses. 
Therefore, we have decided to require LLMs to output in current format.

\subsubsection{Why are text-only instructions not subject to going through the proposed method?}

It is not feasible for two main reasons.
Firstly, the adopted LLMs are not powerful enough to achieve this goal. 
The text-only instruction set is somewhat chaotic, lengthy, diverse in task types, and encompasses various languages. 
For the LLM, simply maintaining the original content is challenging, let alone achieving writing manner alignment.
Secondly, refining the text-only instructions would affect the analysis of the effect of improving the multi-modal instructions, which is the focus of this paper. 
To be honest, the quality of text-only instructions is poor. 
It is no doubt that improving the quality of these text-only instructions can improve the model's performance. 
Given that the difficulty of rewriting text-only instructions, even if the LLM successfully rephrases the text and brings improvement, it is still hard to determine whether this improvement stems from aligning writing manner or eliminating errors.

\subsubsection{Why does QwenVL perform better than LLaVA-7B even though the perplexity of QwenVL is higher than that of LLaVA-7B in Table~\ref{tab:ppl_indicator}?}

The ultimate performance of LMM depends on multiple factors such as parameters, scale and pre-training data, etc, rather than just fitting to instructions. 
Qwen-VL’s pre-training is far more comprehensive that LLaVA’s (1.4B v.s. 558K image-text-pairs in pre-training stage). 
Therefore, comparing the PPL scores across different models and then mapping it to their performance on downstream tasks is unreasonable. 
Moreover, PPL scores are computed with frozen inner LLM, it cannot reflect the fitting ability of LMM with unlocked LLM.

\subsection{Pseudo Code of Writing Manner Alignment}

Algorithm~\ref{alg:instruction alignment} provides a concise pseudocode of our instruction alignment process.

\begin{algorithm}[t]
\caption{Instruction Alignment Pseudocode}
\label{alg:instruction alignment}
\definecolor{codeblue}{rgb}{0.25,0.5,0.5}
\definecolor{codekw}{rgb}{0.85, 0.18, 0.50}
\lstset{
  backgroundcolor=\color{white},
  basicstyle=\fontsize{7.5pt}{7.5pt}\ttfamily\selectfont,
  columns=fullflexible,
  breaklines=true,
  captionpos=b,
  commentstyle=\fontsize{7.5pt}{7.5pt}\color{codeblue},
  keywordstyle=\fontsize{7.5pt}{7.5pt}\color{codekw},
}
\begin{lstlisting}[language=python,showstringspaces=false]
# f: generate rewrite prompt
# g: generate review prompt
# post_process: split answer content from LLM response

for (q, a) in loader: # load a round of conversation
    # Stage 1: LLM Rewrite
    rewrite_prompt = f(q, a)
    rewrite_response = LLM(rewrite_prompt)
    modified_a, status = post_process(rewrite_response)
    if status == False:
        continue
    
    # Stage 2: LLM Review
    review_prompt = g(q, a, modified_a)
    review_response = LLM(review_prompt)
    if "The Revised Answer is fine" in review_response:
        replace(a, modified_a) # replace a with modified_a
\end{lstlisting}
\end{algorithm}

\subsection{Case Study}
\label{sec:case study}

\subsubsection{Positive LLM-aligned Samples.}
\begin{figure*}[!t]
    \centering
    \centering
    \includegraphics[width=15cm]{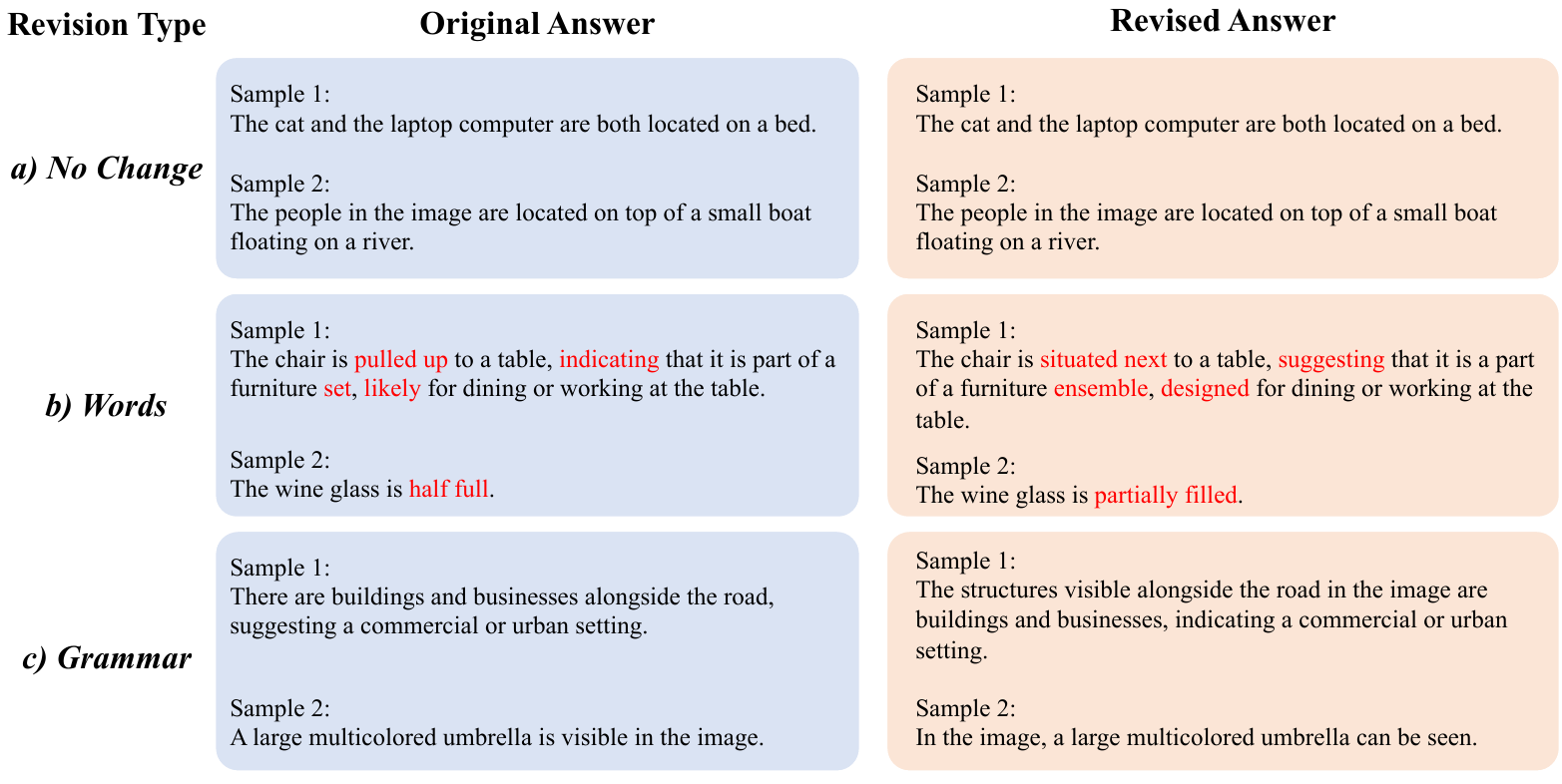}
    \centering
    \includegraphics[width=15cm]{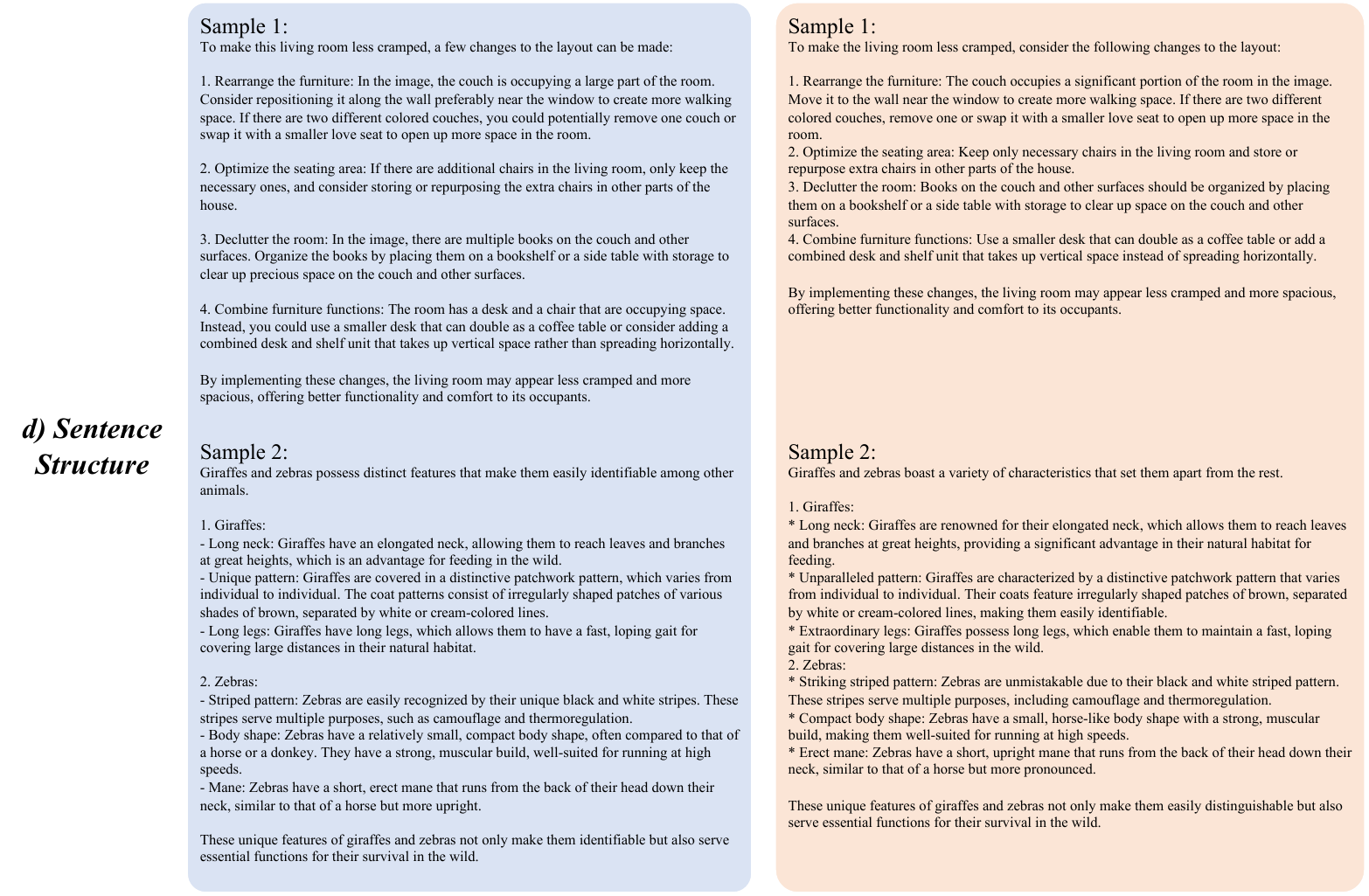}
    \caption{\textbf{Positive LLM-aligned samples in different revision types.} 
    }
    \label{fig:positive_samples}
\end{figure*}

In Figure~\ref{fig:positive_samples}, we showcase several examples of writing manner alignment, primarily categorized into four types: 1) Keep unchanged; 2) Slight adjustment in word choices; 3) Adjustment in grammar; 4) Changes in sentence structure.

\subsubsection{Unqualified LLM-aligned Samples.}
\begin{figure*}[ht]
    \centering

    \centering
    \includegraphics[width=13cm]{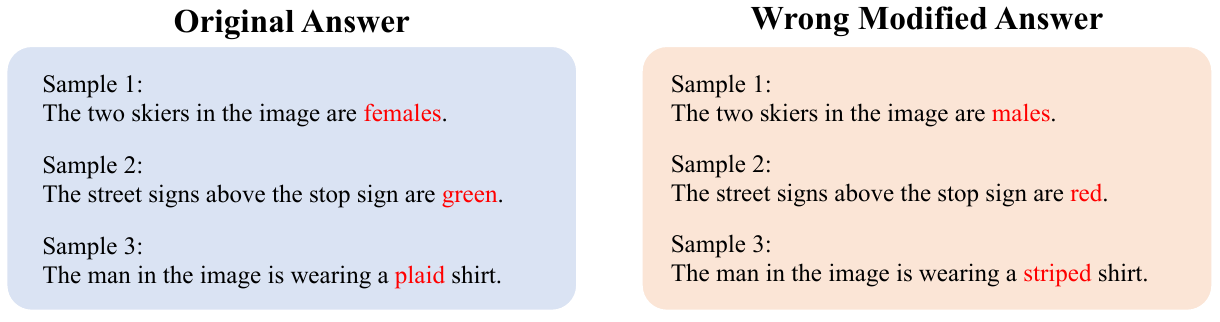}

    \caption{\textbf{Unqualified samples filtered out during the review stage.} 
    }
    \label{fig:unqualified_samples}
\end{figure*}

Figure~\ref{fig:unqualified_samples} describes three samples deemed unqualified during the review stage, showing that the LLM can filter out obvious errors in the rewritten answers.
Combined with the quantitative statistics in Table~\ref{tab:numbers}, the error rate of LLM-based writing manner alignment is low.


\subsubsection{Qualitative Comparisons.}
\label{sec:qualitative}
\begin{figure*}[t!]
    \centering
    \includegraphics[width=13cm]{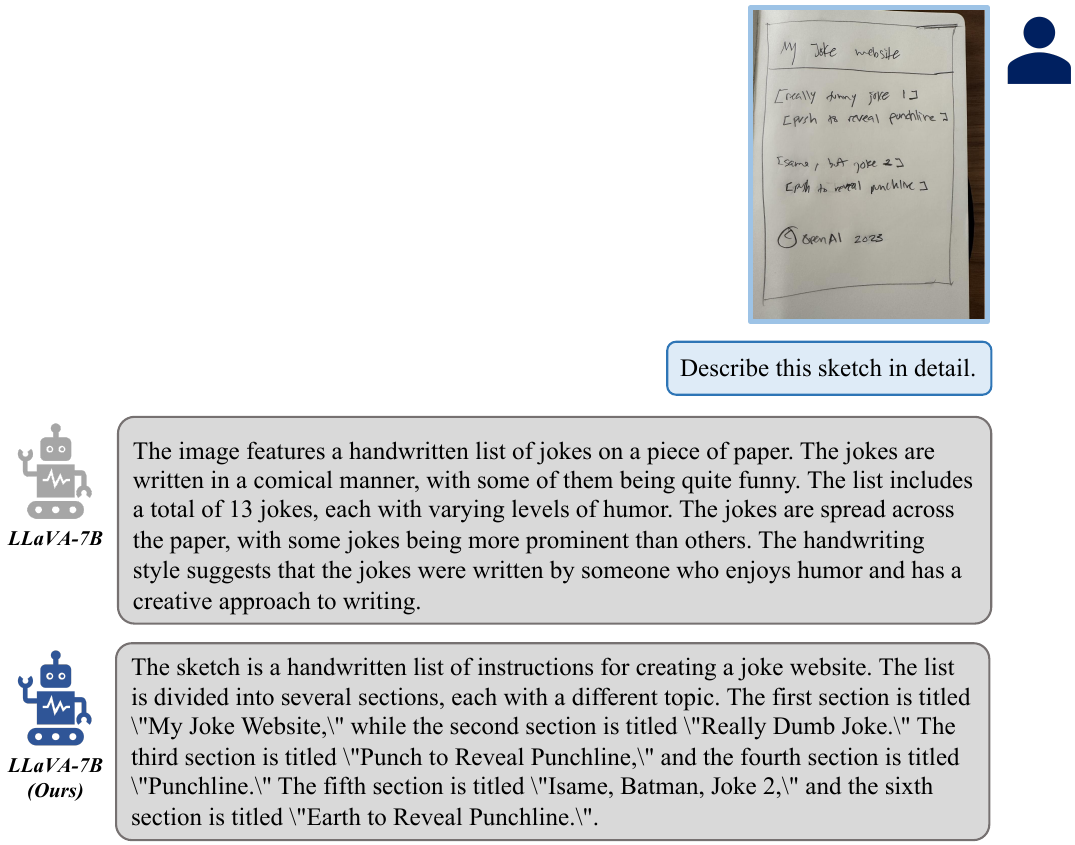}
    \caption{\textbf{Quality comparison in sketch recognition.} Our model performs more accurate understanding in sketch structure. 
    }
    \label{fig:conv_1}
\end{figure*}

\begin{figure*}[t!]
    \centering
    \includegraphics[width=12.8cm]{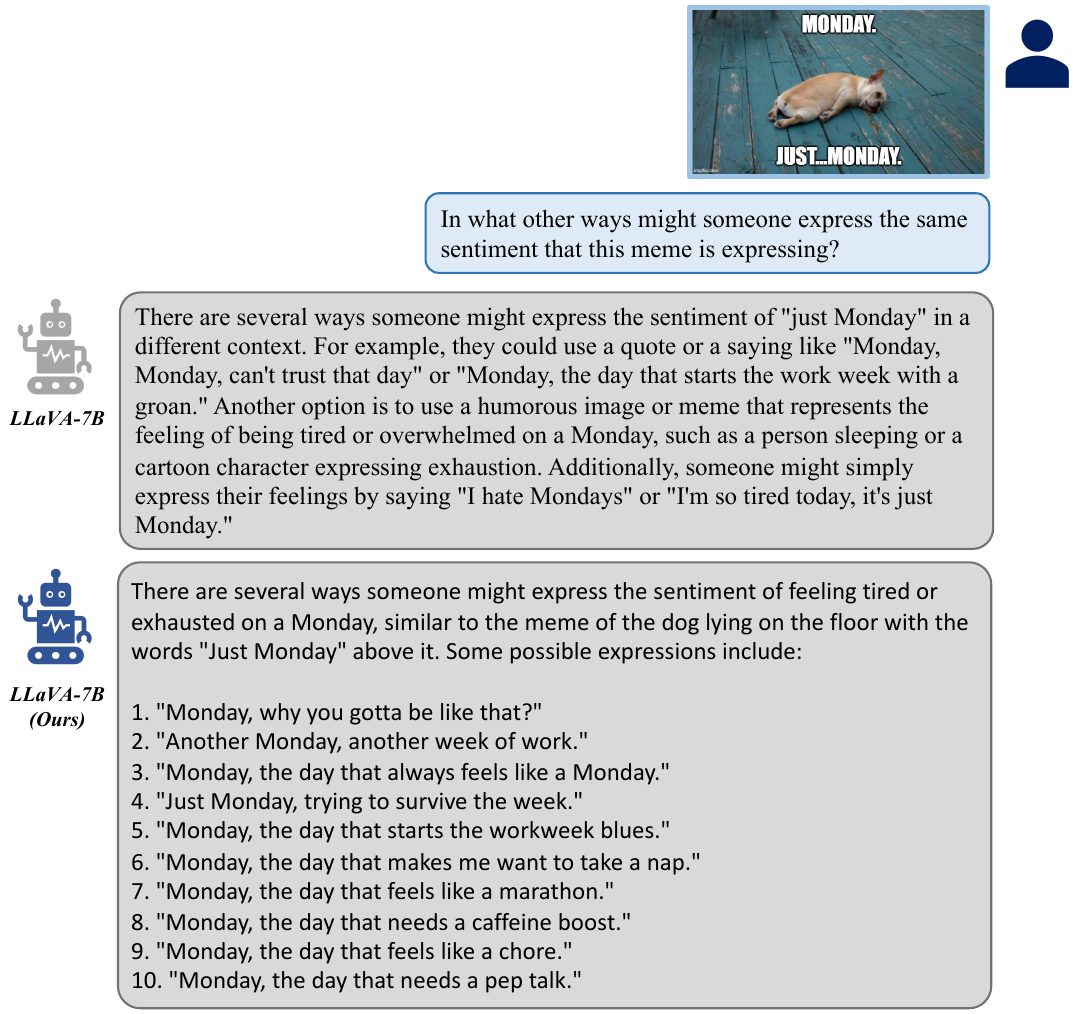}
    \caption{\textbf{Quality comparison in meme recognition.} Our model more closely follows the user's intent, and generates responses in a more clear format.
    }
    \label{fig:conv_2}
\end{figure*}

\begin{figure*}[t!]
    \centering
    \includegraphics[width=12.8cm]{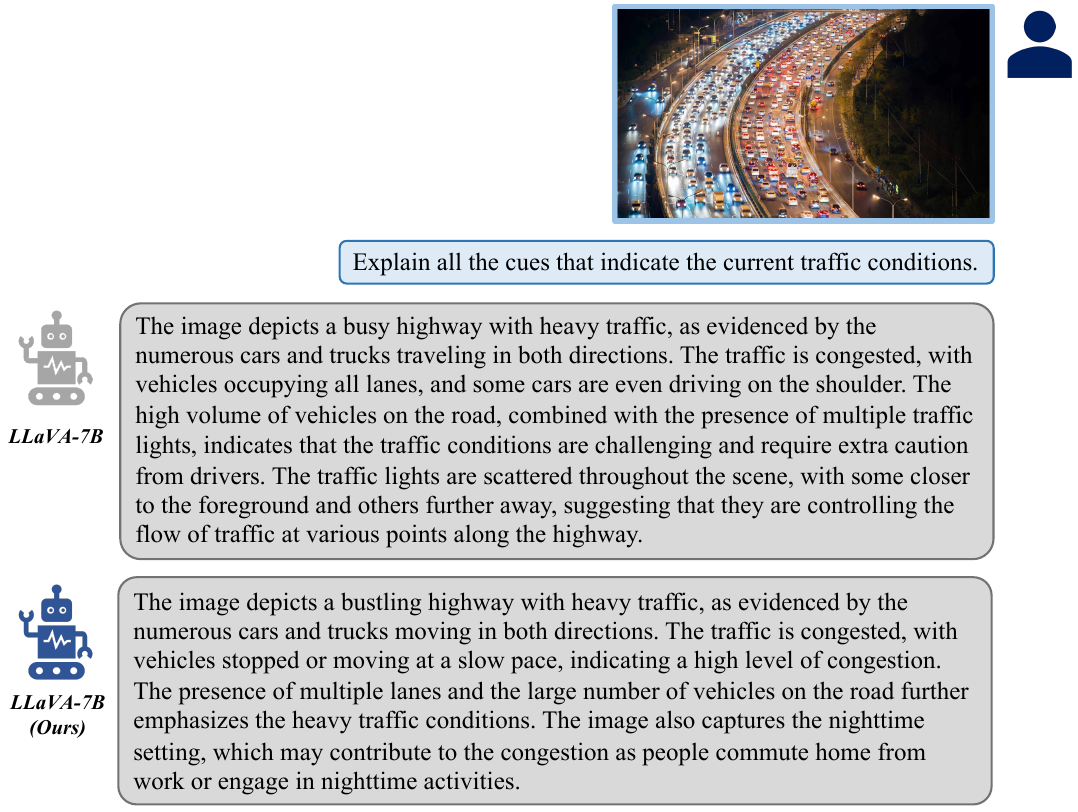}
    \caption{\textbf{Quality comparison in complex scene understanding.} The answer of our model contains fewer visual hallucinations.
    }
    \label{fig:conv_3}
\end{figure*}

In Figure~\ref{fig:conv_1}, Figure~\ref{fig:conv_2}, and Figure~\ref{fig:conv_3}, we present three representative complex visual dialogues.
By comparison, the proposed LLM-aligned trainset enhances LLaVA-7B's capabilities on text recognition, logical reasoning and instruction-following, while also reducing visual hallucinations.

\newpage

\end{document}